\documentclass{article}

\usepackage[final]{neurips_2021}

\makeatletter
\renewcommand{\@noticestring}{%
1st Workshop on Human and Machine Decisions
(WHMD 2021) at NeurIPS 2021.}
\makeatother

\usepackage[utf8]{inputenc}
\usepackage[T1]{fontenc}    
\usepackage{hyperref}       
\usepackage{url}           
\usepackage{booktabs}       
\usepackage{amsfonts}       
\usepackage{nicefrac}       
\usepackage{microtype}     
\usepackage{xcolor}         
\usepackage{amsmath,bm}
\usepackage{graphicx}
\DeclareMathOperator*{\argmax}{arg\,max}

\title{Neural-Symbolic Integration for Interactive Learning and Conceptual Grounding}

\author{
  Benedikt Wagner\\
  Department of Computer Science\\
  City, University of London\\
  London, EC1 0HB, UK \\
  \texttt{Benedikt.Wagner@city.ac.uk}
  \And
  Artur d'Avila Garcez \\
  Department of Computer Science\\
  City, University of London \\
  London, EC1 0HB, UK \\
  \texttt{A.Garcez@city.ac.uk} 
}

\begin{document}

\maketitle

\begin{abstract}
  We propose neural-symbolic integration for abstract concept explanation and interactive learning. Neural-symbolic integration and explanation allow users and domain-experts to learn about the data-driven decision making process of large neural models. The models are queried using a symbolic logic language. Interaction with the user then confirms or rejects a revision of the neural model using logic-based constraints that can be distilled into the model architecture. 
  The approach is illustrated using the Logic Tensor Network framework alongside Concept Activation Vectors and applied to a Convolutional Neural Network.
\end{abstract}

\section{Introduction}
The widespread application of Machine Learning (ML) has caused a surge in interest and research around explainability. Although differing in motive, there is increasing demand for comprehending the processes used by large trained models to arrive at certain decisions, and for evaluating whether such processes conform with human understanding. 
Explanations are offered in a variety of media with varying levels of fidelity and intuition \citep{Confalonieri2021,Guidotti2018} with a substantial part achieved through post-hoc explanation algorithms.   
In most cases, the ML system produces statistical explanations which help determine how the output is derived. 
Most current approaches focus on decoding the behaviour of a system by examining input or output characteristics. However, using background knowledge may provide explanations which are more in line with human conceptualisation, and thus more useful in applications. \\
In the process of providing explanations, one of the limitations of statistical methods is that in general they do not consider domain or background knowledge. 
In light of the above limitations, it may be beneficial to combine statistical techniques with symbolic approaches. We argue that the ease of interpretation of symbolic representations through their interaction with powerful neural network learning algorithms can offer a promising direction for explainability and comprehensibility. \\
\cite{Futia2020} underline the importance of neural-symbolic integration for explainability by suggesting that traditional explainable AI (XAI) methods lack the ability to provide explanations for the variety of target audiences. While most of the explainability methods may be valuable at providing insight to ML experts, domain experts in applications such as finance or healthcare may struggle to interpret the explanations given. 
It is proposed that \emph{interactive integration with semantically-rich representations is key to refining explanations targeted at different stakeholders} \cite{Futia2020}. Indeed, having flexibility in the abstract representation of the information that forms an explanation is key to leveraging domain expertise through interaction and revision of the decision making process.\\
More specifically, neural-symbolic integration would allow us to overcome the static nature of the current ML paradigm. Explainability methods of today do not provide an ability to act on extracted information. Upon finding an undesired property, the only way to influence the model is to retrain it until a good model is found. However, retraining as a process may be unguided and can only be influenced indirectly by the explanation through the collection of additional data.
The result is that many explanation methods become limited in their usefulness, with retraining commonly resulting in catastrophic forgetting of previously acquired information.  
The inclusion of symbolic knowledge during training of deep networks in the form of first-order logic constraints on the loss function has been shown to improve fairness while maintaining the performance of the ML system \citep{Wagner2021} by direct comparison with existing fairness-specific methods using various metrics.\\
We propose to define XAI as the alignment of model behaviour with human values achieved through model comprehensibility and revision.
If one wishes to obtain intuitive, human-like explanations then the alignment must take place at a high level of abstraction with an ability to drill down to deeper explanations as the need arises, as in the case of a child's sequence of \emph{why} questions. While the predominant low-level and statistical explanations are effective for debugging, logical and reasoning-based explanations require a more abstract knowledge representation using higher-level concepts. 
\begin{figure}[!h]
\centering
    \includegraphics[width=0.65\textwidth]{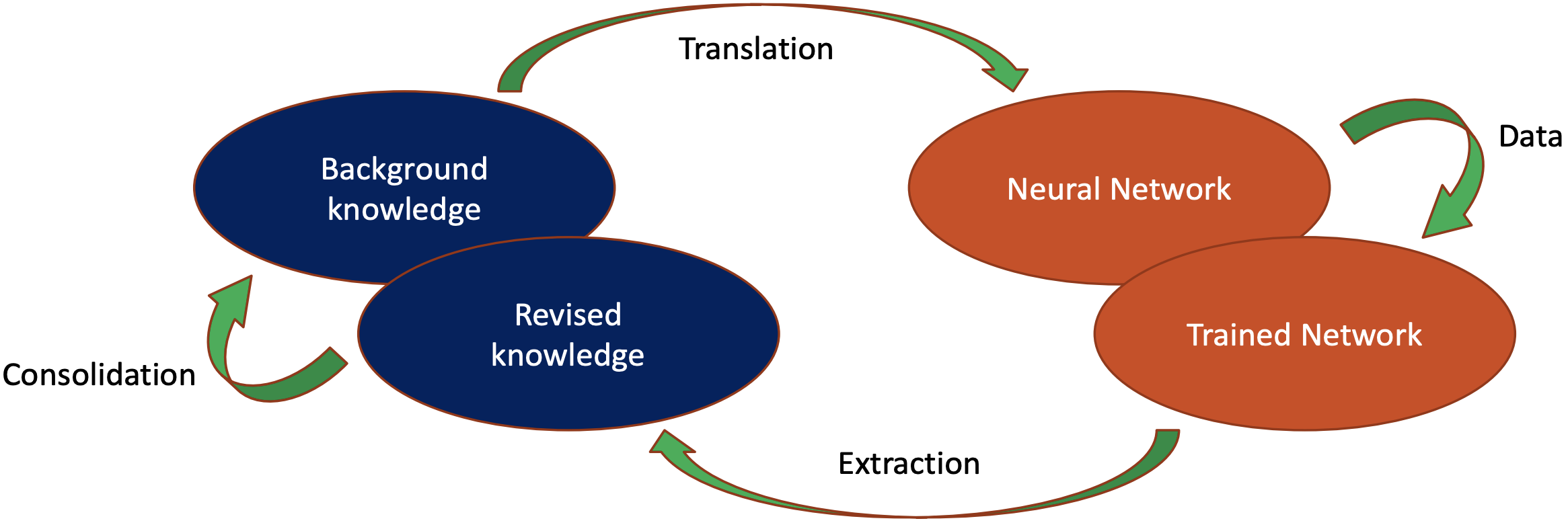}
    \caption{{\small Illustration of the neural-symbolic cycle: knowledge extraction will be carried out by querying a deep network interactively and learning continually, thereby repeatedly applying the neural-symbolic cycle until values are deemed to have been aligned. 
    The neural-symbolic cycle enables a human-in-the-loop approach by offering a common ground for communication and system interaction. Symbolic knowledge representations extracted from the learning system at an adequate level of abstraction for communication with the user will allow for knowledge consolidation and targeted revision of the model.}}
    \label{fig:nesycycle_legacy}
\end{figure}
\section{Interactive Learning and Conceptual Grounding}
Our aim is to query the neural network for symbolic knowledge so that a direct interpretation of abstract representations and operations on those representations become feasible. The approach seeks to explain and possibly revise a decision making process post-hoc and to be model-agnostic. Furthermore, to ensure that the model can adapt to the complexity of tasks in the usual way as popularized by current ML, we aim to retain the advantages of gradient-based, end-to-end learning. As a means of ensuring that the common \emph{communication layer} is not hindered by irreconcilable disparities between the symbolic (discrete) and neural (continuous) representations, we will need to ensure that the model can be queried with human-interpretable operations at an adequate level of abstraction.\\
We use neural-symbolic integration as the bridge at this communication layer: logic provides the semantic precision required of the questions to be put to the model. Although the use of logic may seem to be a barrier at first sight, it is required to formalise knowledge and interaction with a precise semantics. The gap that may exist can in principle be filled by building wrappers to formulate logical queries e.g. using natural language \citep{Singh2020ExploringLogic}. 
The core building block will be the usual logical operators. The goal is to provide an intuition into the operations that the ML model has inferred based on the observed data at a given task. The logical operators \emph{connect} the symbol representations also in the usual way. Symbols are tangible references which will be used to denote abstract concepts that arise through learning of model-specific, data-driven representations. These abstract concepts will be derived from within a trained model, giving rise to explanations that are grounded on the model's inherent representation and operations. \\
The neurosymbolic framework adopted in this paper is that of Logic Tensor Networks (LTN) as described in \cite{Serafini2016} and \cite{Badreddine2020}. However, instead of treating the learning of the parameters from data and knowledge as a single process, we emphasise the dynamic and flexible nature of training from data followed by querying the trained model for knowledge, followed by consolidating that knowledge in the form of constraints for further training, as part of a cycle with stopping criteria defined by the user. We make LTN iterative by saving the parametrization learned at each cycle in our implementation while not requiring the use of any specific model architecture, thus making the approach fully model-agnostic.\\
LTN implements a many-valued first-order logic (FOL) language $\mathcal{L}$ into deep networks.
The syntax of LTN is that of FOL, with formulas consisting of predicate symbols, here denoting concepts, and the connectives: negation ($\neg$), conjunction ($\land$), disjunction ($\lor$), implication ($\rightarrow$), as well as universal ($\forall$) and existential ($\exists$) quantification.\\
To emphasize that symbols are interpreted according to their grounding onto real numbers, LTN uses the term \emph{grounding}, denoted by $\mathcal{G}$, in place of interpretation. Here, we are specifically interested in the grounding of predicates which make up the symbols that refer to the abstract concepts which will form the basis of our model explanation. 
Predicates are grounded as mappings onto the interval $[0,1]$ representing the predicate's degree of truth given the input. 
In order to allow for flexible gradient-based learning with logical constraints, the logical formulas are made differentiable. This is done by defining the connectives according to fuzzy logic connectives approximated by so-called t-norms, t-conorms and fuzzy implication and negation. These are mathematical operations that satisfy a set of logical properties. In the same manner, quantification such as $\forall$ (for all) and aggregation of logical formulas into a set (i.e. a knowledge-base) 
are defined using generalised mean. 
For an extensive explanation of LTN, we refer the reader to \citep{Badreddine2020}.  \\
Our adaptation of the LTN framework can be deployed after training as will be demonstrated. Here, specific outputs and inner representations of any neural network are mapped onto a predicate $P$ to obtain an explanation for $P$ w.r.t. other predicates (other outputs and inner representations) with the use of the logical connectives. Once a relation among the predicates has been established as a logical rule, one can impose additional constraints onto the network with this rule to seek to understand the relationship between this and other possible rules. If we consider all groundings of mappings of a network to be learnable, they will all depend on a set of parameters $\bm{\theta}$. 
The background knowledge consists of a (possibly empty) set of logical rules, referred to as $\phi$, which entails all predicate mappings.
Since the grounding of a rule $\mathcal{G}_\theta(\phi)$ denotes the degree of truth of $\phi$, one natural training signal is the degree of truth of all the rules, including mappings in the knowledge-base $\mathcal{K}$.
The objective function is therefore to maximise the satisfiability of all the rules in $\mathcal{K}$,
$\bm\theta^\ast = \argmax_{\bm\theta\in\bm\Theta}\ \mathrm{Sat}_{A}(\mathcal{G}_\theta(\mathcal{K}))$,
which is subject to an aggregation $A$ of the rules in $\mathcal{K}$. 

\begin{figure}[!h]
    \centering
    \includegraphics[width=.87\textwidth]{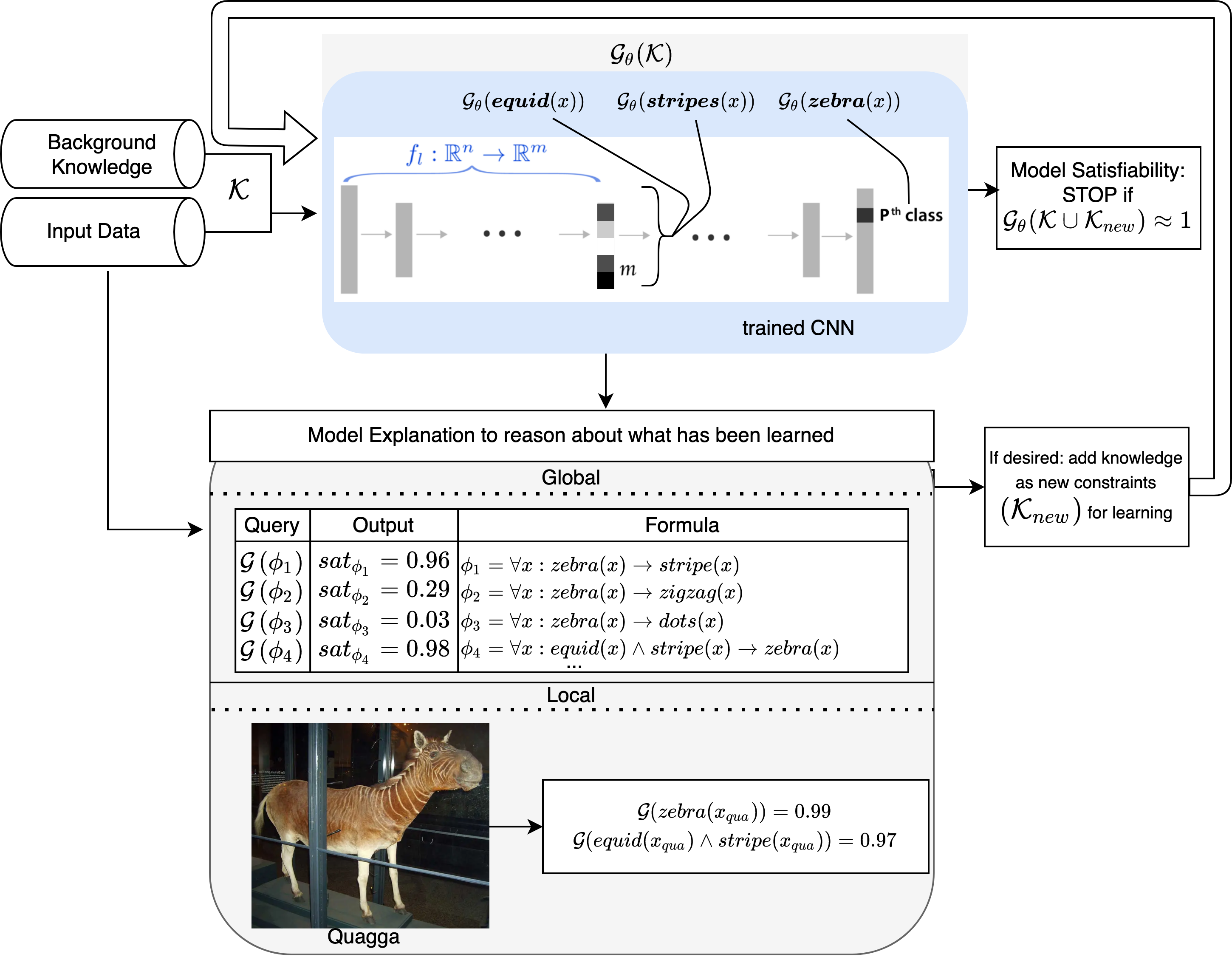}
    \caption{{\small We produce local explanations (for individual inputs/images) and global explanations (universally-quantified formulas) for the deep learning model by querying specific neurons. We then reason about the generality of the explanations given the \textit{truth queries} extracted from the trained network. The figure shows some of these queries associated with groundings in the neural network and their satisfiability (sat) levels. Using linear probes to ground the activation patterns of intermediate representations into the language of LTN, we are able to utilise abstract concepts as explanation symbols in the logic. Following querying, the neural model can be constrained based on a user selection of logical formulas $\mathcal{K}_{new}$ for further learning. This iterative process seeks to align the model with user values in the context of background knowledge $\mathcal{K}$. In the figure, the Quagga is classified as a zebra. A user's desire to change such classification should trigger the addition of knowledge into $\mathcal{K}_{new}$ informed by the queries to be satisfied by the final trained model. Notice that training from data may begin without any background knowledge which can be revised by querying user-defined concepts and constraints deemed as necessary for the network to learn.}  
    }
    \label{fig:quagexpl}
\end{figure} 
\section{Demonstration of Grounding for Concept Interaction Explanation}
To obtain conceptual explanations that provide comprehensible descriptions of what has been learned, we must ground low-level information into reusable concepts that are present at hidden representations within the network. 
Let us illustrate the idea with an example which we have implemented in LTN to explain a Convolutional Neural Network (CNN). We draw inspiration from the TCAV approach \citep{Kim2017} but modify it substantially for the implementation in LTN. Consider any neural network that takes as input $\bm{x} \in \mathbb{R}^{n}$, which projects onto any layer $l$ within the network consisting of $m$ neurons, according to a function: 
$f_{l} : \mathbb{R}^{n} \rightarrow \mathbb{R}^{m}$. In an iterative explanation process, we seek to connect representations inside the network. There is no restriction on which layer $l$ to use, but in a CNN this is generally the layer immediately before the fully-connected layer (i.e. the classifier) \citep{Odense2020}. 
We adapt TCAV \cite{Kim2017} to enable users to specify the concepts to be checked (queried) at any adequate level of abstraction. \cite{Graziani2018} have shown in an application to medical imaging that domain-related concepts can be valuable for gaining insight into the decision making. The approach proposed in this paper extends this idea to allow for complex concept interactions (as defined by the logic) and model retraining using such logical rules as constraints. Using random examples alongside a user-defined set of examples (images) that capture a concept, we form a linear probe at layer $l$ which evaluates the activation values produced by the examples and already known concepts from which further data can be selected for use. 
In \cite{Kim2017}, the linear probe at layer $l$ serves as a building block for the Concept Activation Vector used to calculate conceptual sensitivities of inputs and classes. In this paper, we integrate the concept mapping directly into the interactive framework, allowing concepts to be combined into the logic, evaluated using fuzzy logic, and chosen for further training of the neural network model.
This linear probe works as a classifier for our conceptual grounding, which is then integrated as a logical predicate into the LTN framework. The top part of Figure \ref{fig:quagexpl} illustrates the process. At each time that a user wishes to distil a model inference into specific concepts $C$, they simply need to select a set $P_C$ of positive examples and a set $N$ of negative examples. The linear probe then serves to distinguish the activation values of the neurons in layer $l$ between $\left\{f_{l}(\boldsymbol{x}): \boldsymbol{x} \in P_{C}\right\} $ and $\left\{f_{l}(\boldsymbol{y}): \boldsymbol{y} \in N\right\}$. 
This has the advantage of not being bound by pre-existing data or features. Independent of the original task, examples (images) may be collected and any number of user-defined concepts checked (queried) against the network.\\ 
As an example, we query a GoogLeNet model \citep{Szegedy2015} trained on ImageNet to explain the output class of zebras with respect to user-defined concepts, as illustrated in \cite{Kim2017} and the bottom part of Figure \ref{fig:quagexpl}. We extract four different concept descriptions using images from the Broden dataset \citep{Bau2017} to dissect the \emph{zebra} classification into the concepts of \emph{stripes}, \emph{dots}, \emph{zigzags} and an abstract representation of the horse-family concept \emph{equidae} (horses, donkeys, zebras and others). We learn the groundings of the activation patterns for the specified concepts from 150 images of each concept and an equal number of negative examples for each class. Subsequently, the truth-value of each query is calculated through fuzzy logic inference using LTN. These queries can be specific to an image (local) or aggregated across the entire set of examples (global). \\ 
The quantifier $\forall$ is used to aggregate across a set of data points by replacing $x$ with every image available from the dataset thus evaluating the model's behavior across all available data. 
The following implication: $\forall x: zebra(x) \rightarrow stripe(x)$, with the symbol $stripe(x)$ being replaced by the corresponding concept grounding in the network, provides an insight onto how important the concept $stripe(x)$ is for the CNN's classification output $zebra(x)$ given the set of images $x$. \\
Furthermore, we can combine several concepts using the logic:
$\forall x: equid(x) \land stripe(x) \rightarrow zebra(x)$ returns a truth value of $0.98$ across a set of 3000 examples from ImageNet, indicating that the CNN assigns any horse-like object with stripes to the class of zebras. 
When applying a universal quantifier, the user is able to evaluate the decision making process of the model in general, by examining the concepts on all available data, even if it has not been used for training, thereby producing a global explanation.\\
One example previously unknown to the model is the extinct quagga, an animal characterised by a brown striped coat instead of the black and white pattern of zebras which has been selected to illustrate the potential of local explanations. The model identifies this animal correctly as a member of the Equidae family, recognises the stripes on the animal and consequently classifies the image as that of a zebra, as shown in Figure \ref{fig:quagexpl}.
By utilising the trained linear probes of the activation vectors to ground individual images, we generate local explanations that provide insight into why a particular image might be classified in a certain way according to the model.

Upon identifying potential undesired behaviour, for example by querying known exceptions, a user can add new rules into the knowledge-base (by adding logical formulas into $\mathcal{K}_{new}$) for further training of the network.
In case the specification of quagga and zebra is to be changed, an alternative inference process can be imposed on the CNN model. Recall that quagga are currently considered by the CNN to be a subspecies of zebra. Assuming that the user decides to change this, as an example, let us consider introducing concept probes $bw(x)$ for \emph{black and white} objects and $col(x)$ for \emph{colourful objects}, and let us assume that these concepts are to be regarded as mutually exclusive. Adding the following rule to $\mathcal{K}_{new}$ as a new constraint to be satisfied by learning should force the neural model to only classify black and white objects as zebras: $\phi_5 = \forall x: equid(x) \land stripe(x) \land \neg bw(x) \rightarrow \neg zebra(x)$.\footnote{Notice that the satisfiability of this rule should be the same as that of $\forall x: equid(x) \land stripe(x) \land col(x) \rightarrow \neg zebra(x)$, as we apply a $softmax$ function to mutually exclusive concept probes; in this case $\forall x: col(x) \leftrightarrow \neg bw(x)$.} \\
Before further training, $\phi_5$ exhibits a low $sat$-level of $sat_{\phi_5}=0.09$, as the model classifies all objects associated with the $equid(x)$ and the $stripe(x)$ concepts to the $zebra$ class regardless of their color. By retraining for only five iterations, the $sat$-level increases to $sat_{\phi_5}=0.94$, which indicates that only black and white objects (in conjunction with the \emph{stripe} and \emph{equidae} concepts) are now considered to be zebras. Therefore, the example image of the quagga is no longer inferred to be in the zebra class with $\mathcal{G}(zebra(x_{qua}))=0.08$, where $x_{qua}$ denotes the image of a quagga. The neural model nevertheless identifies correctly the equidae and stripe concepts in the quagga, with  $\mathcal{G}(equid(x_{qua}) \land stripe(x_{qua}))=0.97$.  

\section{Conclusion}
We proposed a method combining TCAV and LTN that allows for complex conceptual explanation and interactive learning. It enables domain experts to learn about the data-inferred decision making process of large ML models by querying the model and defining constraints for further learning as part of an iterative process. We have implemented the idea in LTN and used GoogLeNet model to illustrate its use. \\
Future work will explore alternative model architectures to improve the effectiveness of concept groundings. In particular, vision transformers are capable of providing rich representations that may prove useful.
Furthermore, we shall investigate an alternative to the TCAV approach of providing examples of concepts. Despite the fact that this may be useful in that it allows for customizable concepts, it may be impractical at times to collect data. 
It should be possible to accomplish the same goal by integrating multimodal (vision and language) models that permit natural language description from zero-shot classification.
\bibliography{references.bib}

\end{document}